\documentclass[letter, 10 pt, conference]{ieeeconf}  
\IEEEoverridecommandlockouts                              
\usepackage{blindtext}
\usepackage{graphics}    
\usepackage{times}       
\usepackage{amsmath}     
\usepackage{amssymb}     
\usepackage{graphicx}
\usepackage{subcaption}
\usepackage{xspace}
\usepackage{xcolor}
\usepackage{color}
\usepackage{todonotes}
\usepackage[linesnumbered, ruled]{algorithm2e}
\usepackage{multirow}
\usepackage{balance}
\usepackage{makecell}
\usepackage{url}
\usepackage{fancyhdr}

\usepackage[font=small]{caption}

\def\figref#1{Fig.~\ref{#1}}

\def\eqref#1{~(\ref{#1})}

\title{\LARGE \bf Joint Vision-Based Navigation, Control and Obstacle Avoidance for UAVs in Dynamic Environments}

\author{Ciro Potena$^1$ \and Daniele Nardi$^1$ \and Alberto Pretto$^2$ 
  \thanks{$^{1}$Potena and Nardi are with the Department of Computer, Control, and Management Engineering ``Antonio Ruberti``, Sapienza University of Rome, Italy. Email: { \{potena, nardi\}@diag.uniroma1.it}. 
  $^{2}$Pretto is with IT+Robotics S.r.l., Vicenza, Italy. Email: { alberto.pretto@it-robotics.eu} \newline 978-1-7281-3605-9/19/\$31.00 \textcopyright 2019 IEEE}
}

\begin{document}

\maketitle

\fancyhead[C]{\vspace{-1cm}This paper has been accepted for publication in the 2019 European Conference on Mobile Robots (ECMR)\\DOI: 10.1109/ECMR.2019.8870944 \url{https://ieeexplore.ieee.org/document/8870944}}
\fancyfoot[C]{Please cite this paper as: C. Potena, D. Nardi, and A. Pretto,\\``Joint Vision-Based Navigation, Control and Obstacle Avoidance for UAVs in Dynamic Environments'',\\in Proc. of the European Conference on Mobile Robots (ECMR), 2019.}
\thispagestyle{fancy}
\pagestyle{empty}

\begin{abstract}

This work addresses the problem of coupling vision-based navigation systems for Unmanned Aerial Vehicles (UAVs) with robust obstacle avoidance capabilities. The former problem is solved by maximizing the visibility of the points of interest, while the latter is modeled by means of ellipsoidal repulsive areas.   
The whole problem is transcribed into an Optimal Control Problem (OCP), and solved in a few milliseconds by leveraging state-of-the-art numerical optimization.  
The resulting trajectories are well suited for reaching the specified goal location while avoiding obstacles with a safety margin and minimizing the probability of losing the route with the target of interest.
Combining this technique with a proper ellipsoid shaping (i.e., by augmenting the shape proportionally with the obstacle velocity or with the obstacle detection uncertainties) results in a robust obstacle avoidance behavior.
We validate our approach within extensive simulated experiments that show effective capabilities to satisfy all the constraints even in challenging conditions. We release with this paper the open source implementation.

\end{abstract}

\section{Introduction}
\label{sec:intro}

In recent years, small unmanned aerial vehicles (UAVs) have increasingly gained popularity in many practical applications thanks to their effective survey capabilities and limited cost. For some basic applications, solutions are already available on the market to localize the drone and provide some basic navigation and obstacle avoidance capabilities. However, to safely navigate in the presence of obstacles, an effective and reactive planning algorithm is an essential requirement. 

On the other hand, thanks to the progress in perception and control algorithms \cite{greeff2018,neunert2016fast}, and to the increased computational capabilities of embedded computers, vision-based optimal control techniques became a standard for UAVs moving in dynamic environments \cite{Falanga2018,Sheckells2016}. They allow to mitigate some of the vision-based perception limitations (e.g. feature tracking failures) through an ad-hoc trajectory planning and have partially solved the vehicle state-estimation problem that, in the last decades, has been commonly faced with motion capture systems.

\begin{figure}
  \centering
  \includegraphics[width=0.95\columnwidth]{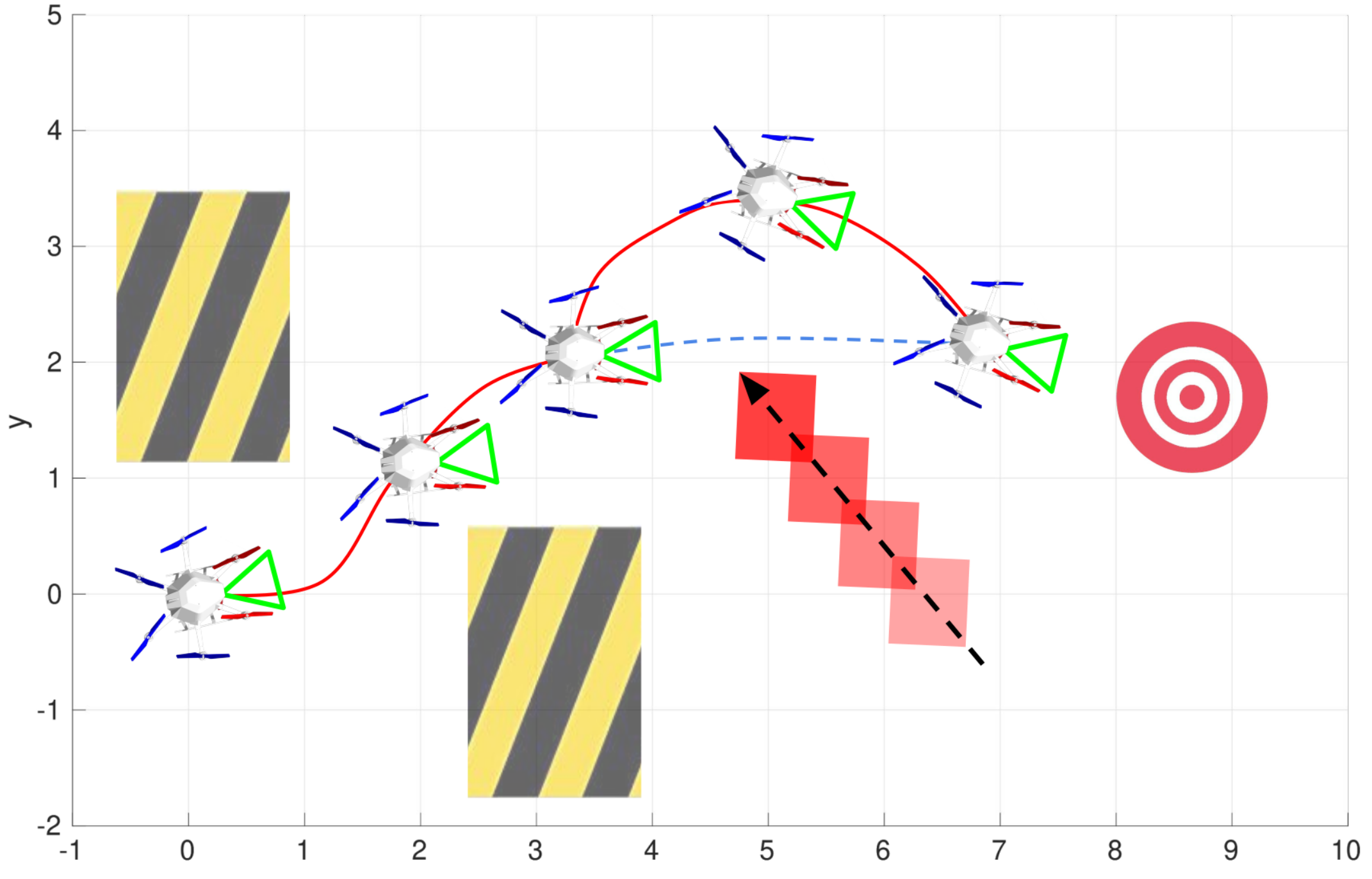}
  \caption{An example of application for the proposed system: an UAV is asked to reach a desired state while constantly framing a specific target (the red circular target in the picture). The environment is populated with static obstacles (black and yellow striped objects in the picture) that should be avoided. Dynamic obstacles (e.g., other agents, depicted with the red moving box) may suddenly appear and block the planned trajectory (i.e., the blue dashed line). With the proposed method, the UAV reacts to the detected object by steering along a new safe trajectory (i.e., the red line).}
  \label{fig:teaser}
\end{figure}

However, the problem of addressing perception and obstacle avoidance together has been rarely investigated \cite{Falanga2017}. In this article, we take a small step forward in this direction by proposing an optimal controller that takes into account in a joint manner the perception, the dynamic, and the avoidance constraints (\figref{fig:teaser}).
The proposed system models vehicle dynamics, perception targets and obstacles in terms of Non Linear Model Predictive Controller (NMPC) constraints: dynamics are accounted by providing a non-linear dynamic model of the vehicle, while targets are modeled by targets visibility constraints in the camera image plane and obstacles by repulsive ellipsoidal areas, respectively. The proposed system also allows incorporating estimation uncertainties and obstacle velocities in the ellipsoids, allowing to deal also with dynamic obstacles. 

The entire problem is then transcribed into an Optimal Control Problem (OCP) and solved in a receding horizon fashion: at each control loop, the NMPC provides a feasible solution to the OCP and only the first input of the provided optimal trajectory is actually applied to control the robot. By leveraging state-of-the-art numerical optimization, the OCP is solved in a few milliseconds making it possible to control the vehicle in real-time and to guarantee enough reactivity to re-plan the trajectory when new obstacles are detected.

Moreover, our approach does not depend on a specific application and can potentially provide benefits to a large variety of applications, such as vision-based navigation, target tracking, and visual servoing.

We validate our system through extensive experiments in a simulated environment. We provide an open source C++ implementation of the proposed solution at:
 
 \begin{quote}
   \begin{footnotesize}
     { \url{https://github.com/cirpote/rvb_mpc}}
   \end{footnotesize}
\end{quote}


\subsection{Related Work}
\label{sec:related}

A vision-based UAV needs three main components to effectively navigate in a dynamic environment:
\begin{enumerate}
 \item A reactive control strategy, to accurately track a desired trajectory while reducing the motors effort; 
 \item A reliable collision avoidance module, to safely explore the environment even in presence of dynamic, unmodeled obstacles;
 \item An adaptive, perception aware, on-line planner, to support the vision-based state estimation or to constantly keep the line-of-sight with a possible reference target.
\end{enumerate}

A wide literature addressing individually, or in pairs, these requirements is available, among others: \cite{potena2017ecmr, kamel2017, Usenko2017, baca2018jfr, thomas2016}. However, they have rarely been addressed together, in particular when dealing with unexpected and moving obstacles.\\
The requirement 1) is an essential capability for highly dynamic vehicles such as UAVs, hence extensively covered in literature, and often formulated as an OCP \cite{KirkOCP}. Model Predictive Controller (MPCs) is a well-known control technique capable to deal with OCPs, and have recently gained great popularity thanks to increased on-board computational capabilities of embedded computers. In \cite{houska2011ocam} and \cite{quirynen2015ocam} ACADO, a framework for fast Nonlinear Model Predictive Control (NMPC), is presented; \cite{kamel2015cca} and \cite{potena2017ecmr} use ACADO for fast attitude control of UAVs. 
In our previous work \cite{pdc2018}, we proposed a solution to improve the real-time capabilities of NMPCs. We aided the controller with a time-mesh strategy that refines the initial part of the horizon inside a flat output formulation. In \cite{greeff2018}, the authors addressed the reactive control problem by building upon a flatness-based Model Predictive Control: the approach converts the optimal control problem in a linear convex quadratic program by accounting for the non-linearity in the model through the use of an inverse term. Experiments performed in simulation and real environments demonstrate improved trajectory tracking performance. In \cite{neunert2016fast}, the authors propose to employ an iterative optimal control algorithm, called Sequential Linear Quadratic, applied inside a Model Predictive Control setting to directly control the UAV actuation system.\\

The collision-free trajectory generation (requirement 2) is usually categorized into three main strategies: search-based approaches \cite{jung2013,Richter2016}, optimization-based approaches \cite{Usenko2017,Oleynikova2016}, path sampling and motion primitives \cite{mueller2013,Paranjape2013}.

In \cite{Oleynikova2016} the authors propose a motion planning approach able to run in real-time and to continuously recompute safe trajectories as the robot perceives the surrounding environment. Although the proposed method allows to replan at a high rate and react to previously unknow obstacles, it might be vulnerable to vision-based perception limitations.\\

Steering a robot to its desired state by using visual feedback obtained from one or more cameras (requirement 3) is formally defined as Visual Servoing (VS), with several applications within the UAV domain \cite{pestana2014,mebarki2015,lee2012,Falanga2017}. Among the others, in Falanga \textit{et al.}~\cite{Falanga2017}, the authors address the flight through narrow gaps by proposing an active-vision approach and by relying only on onboard sensing and computing. The system is capable to provide an accurate trajectory while simultaneously estimating the UAV's position by detecting the gap in the camera images. Nevertheless, it might fail in presence of unmodeled obstacles along the path.\\

A fully autonomous UAV navigating in a cluttered and dynamic environment should be able to concurrently solve \textit{all} the three problems listed above. A solution could be to combine three of the methods presented above, to deal with each problem individually. Unfortunately, due to poor integration between methods and the overall computational load, this solution is not easily feasible. Jointly addressing a subset of these problem is a topic that is recently gathering great attention: in \cite{Sheckells2016}, the authors propose to encode in the NMPC cost function the image feature tracks, implicitly keeping them in the field of view while reaching the desired pose. 
Similarly, in our previous work \cite{potena2017ecmr}, we propose a two-steps NMPC.
In Falanga \textit{et al.} \cite{Falanga2018}, the authors propose a different version of NMPC that also takes into account the features velocity in the camera image plane. The controller will eventually steer the vehicle keeping the features as close as possible to the image plane center, while minimizing their motion. This mitigates the blur of the image due to the camera motion, aiding the target detection and the features tracking. However, the methods presented so far in general do not guarantee a fully autonomous flight in cluttered environments or in presence of unmodeled obstacles.\\

In \cite{garimella2017}, the authors propose a NMPC which incorporate obstacles in the cost function. To increase the robustness in avoiding the obstacles, the UAV trajectories are computed taking into account the uncertainties of the vehicle state. Kamel \textit{et al.} \cite{kamel2017} deal with the problem of multi-UAV reactive collision avoidance. They employ a model-based controller to simultaneously track a reference trajectory and avoid collisions. The proposed method also takes into account the uncertainty of the state estimator and of the position and velocity of the other agents, achieving a higer degree of robustness. Both these methods show a reactive control strategy, but might not allow the vehicle to perform a vison-based navigation.

\subsection{Contributions}
\label{sec:contribution}

Our contributions are the following: (i) an optimal control method that incorporates simultaneously both perception and obstacle avoidance constraints; (ii) a flexible obstacle parameterization that allows to model different obstacle shapes and to encode both obstacles' uncertainty and speed; (iii) an open-source implementation of our method. Our claims are backed up through the experimental evaluation.

\section{Problem Formulation}
\label{sec:approach}

The goal of the proposed approach is to generate an optimal trajectory that takes into account perception and action constraints of a small UAV and, at the same time, allows to safely fly through the environment by avoiding all the obstacles that can possibly lie along the planned trajectory. 
The need to couple action and perception derives from different factors. On the one hand, there are  the vision based navigation limits where, to guarantee an accurate and robust state estimation, it is necessary to extract meaningful information from the image. On the other hand, in some specific cases (e.g. Visual Servoing) the feedback information used to control the vehicle is extracted from a vision sensor, thus the vision target should be kept in the camera image plane. Similarly, taking into account the surrounding obstacles is important to ensure a safe flight in cluttered environments. 
Considering all those factors together allows to fully leverage the agility of UAVs and to have a fully autonomous flight. Therefore, it is essential to jointly consider all these constraints.

Let $l$ be the state vector of the target object (e.g., the 3D point representing the center of mass of a target object), while let $x$ and $u$ be the state and the input vectors of a robot, respectively. Furthermore, let $o$ the state vector of the obstacles to avoid. Let assume the robot's dynamic can be modeled by a general, non linear, differential equations system $\dot{x} = f(x,u)$. Finally, given some flight objectives, we can define an action cost $c_a(x_t,u_t)$, a perception cost $c_p(x_t,l_t,u_t)$, a navigation cost $c_n(x_t,u_t)$, and an avoidance cost $c_o(x_t,o_t,u_t)$
We can thus formulate the coupling of action, perception, and avoidance as an optimization problem with cost function:
\begin{align}
    \label{eq:cost_function}
    {} J = c_f(x_{tf}) + \int_{t_0}^{t_f}c_a(x_t,u_t)+c_p&(x_t,l_t,u_t)+\nonumber\\
    \hspace{0.5cm}c_n(x_t,u_t)+c_o&(x_t,o_t,u_t)dt\nonumber \\  \\
    {} \text{subject to:}\hspace{3cm}&\dot{x}=f(x,u)\nonumber\\
    {} h(x_t,l_t&,o_t,u_t) \leq 0 \nonumber
\end{align}
where $h(x_t,l_t,o_t,u_t)$ stands for the set of inequality constraints to satisfy along the trajectory, $c_f(x_{tf})$ stands for the cost on the final state, and $t_f-t_0$ represents the time horizon in which we want to find the solution. In the following we describe how we model all the cost function components.
\begin{figure}
 \centering
 \includegraphics[width=\columnwidth]{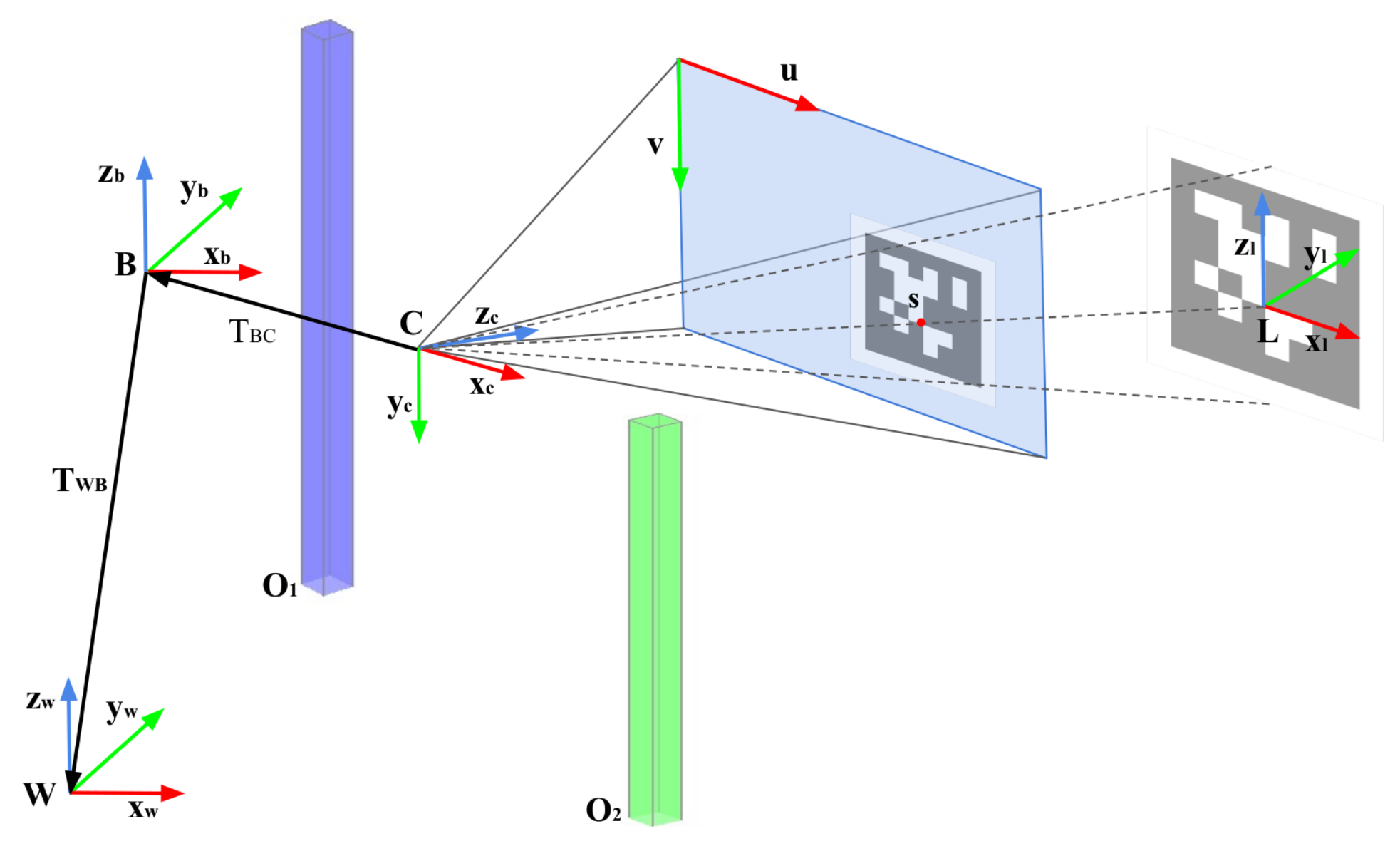}
 \caption{Overview of the reference systems used in this work: the world frame $W$, the body frame $B$, the camera frame $C$, the landmark and obstacles frames $L$ and $O_i$. $T_{WC}$ and $T_{BC}$ represent the body pose in the world frame $W$ and the transformation between the body and the camera frames $C$, respectively. Finally, $s$ is the landmark reprojection onto the camera image plane.}
  \label{fig:ref_systems}
 \end{figure}

\subsection{Quadrotor Dynamics}
\label{sec:dynamics}

In this work, we make use of five reference frames: (i) the world reference frame $W$; (ii) the body reference frame $B$ of the UAV; (iii) the camera reference frame $C$; (iv) the i-th obstacle reference frame $Oi$ and the target reference frame $L$. An overview about the reference systems is illustrated in \figref{fig:ref_systems}.
To represent a vector, or a transformation matrix, we make use of a prefix that indicates the reference frames in which the quantity is expressed. For example, $x_{WB}$ denotes the position vector of the body $B$ frame with respect to the world frame $W$, expressed in the world frame. 

According to this representation, let $p_{WB} = (p_x, p_y, p_z)^T$ and $r_{WB}=(\phi, \theta, \psi)^T$ be the position and the orientation of the body frame with respect to the world frame, expressed in the world frame, respectively. Additionally, let $V_{WB} = (v_x, v_y, v_z)^T$ be the velocity of the body, expressed in the world frame. Finally, let $u = (T, \phi_{cmd}, \theta_{cmd}, \dot{\psi}_{cmd})^T$ be the input vector, where $T = (0, 0, t)^T$ is the thrust vector normalized by the mass of the vehicle, and $\phi_{cmd}$, $\theta_{cmd}$, $\dot{\psi}_{cmd}$ are the roll, pitch, and yaw rate commands, respectively. Thus, the quadrotor dynamic model $f(x,u)$ can be expressed as:
\begin{align}
    \label{eq:dyn_model}
    v_{WB} &= \dot{p}_{WB} \nonumber\\
    \dot{v}_{WB} =& g_W + R_{WB}T \nonumber\\
    \dot{\phi} = \frac{1}{\tau_{\phi}}(&k_{\phi}\phi_{cmd} - \phi) \\  
    \dot{\theta} = \frac{1}{\tau_{\theta}}(&k_{\theta}\theta_{cmd} - \theta)  \nonumber\\
    \dot{\psi} &= \dot{\psi}_{cmd} \nonumber
\end{align}
where $R_{WB}$ is the rotation matrix that maps the mass-normalized thrust vector $T$ in the world frame, and $g_W = (0, 0, -g)^T$ is the gravity vector. For the attitude dynamics we make use of a low-level controller that maps the high-level attitude control inputs into propellers’ velocity. The $\tau_i$ and $k_i$ parameters are obtained through an identification procedure \cite{kamel2016}.

\subsection{Perception Objectives}
\label{sec:perc_obj}

Let $p_{WL}=(l_x, l_y, l_z)^T$ the 3D position of the target of interest in the world frame $W$. We assume the UAV to be equipped with a camera having extrinsic parameters described by a constant rigid body transformation $T_{BC} = (p_{BC}, R_{BC})$, where $p_{BC}$ and $R_{BC}$ are the position and the orientation, expressed as a rotation matrix, of the camera frame $C$ with respect to $B$. The target 3D position in the camera frame $C$ is given by:
\begin{align}
 p_{CL} = (R_{WB}R_{BC})^T( p_{WL} - (R_{WB}p_{BC} + p_{WB}) )
\end{align}
The 3D point $p_{CL}$ is then projected onto the image plane coordinates $s=(u,v)^T$ according to the standard pinhole model:
\begin{align}
  u = f_x\frac{p_{CLx}}{p_{CLz}},\hspace{1cm} v = f_y\frac{p_{CLy}}{p_{CLz}}
\end{align}
where $f_x$ and $f_y$ stand for the focal lenghts of the camera. It is noteworthy to highlight that we are not using the optical centers parameters $c_x$ and $c_y$ in projecting the target, since it is convenient to refer it with respect to the center of the image plane. 

To ensure a robust perception, the projection $s$ of a target of interest should be kept as close as possible to the center of the camera image plane. Therefore we formulate the perception cost $c_p(x_t,l_t,u_t)$ as:
\begin{align}
  c_p(x_t,l_t,u_t) = s H s^T, \hspace{.5cm} H = h\begin{bmatrix} 
    \frac{1}{f_c}&0\\0&\frac{1}{f_r} \end{bmatrix} 
\end{align}
where $f_c$ and $f_r$ represent the number of columns and rows in the camera image, while $h$ is a weighting factor. With this choice, we penalize more the reprojection error of $s$ in the shorter image axis. For instance, if the camera streams a 16:9 image, the optimal solution will care more to keep $s$ closer to the center of the image along the $v$-axis. 

\subsection{Avoidance and Navigation Objectives}
\label{sec:avoid_obj}

\begin{figure}[t]
 \centering
 \includegraphics[width=0.8\columnwidth]{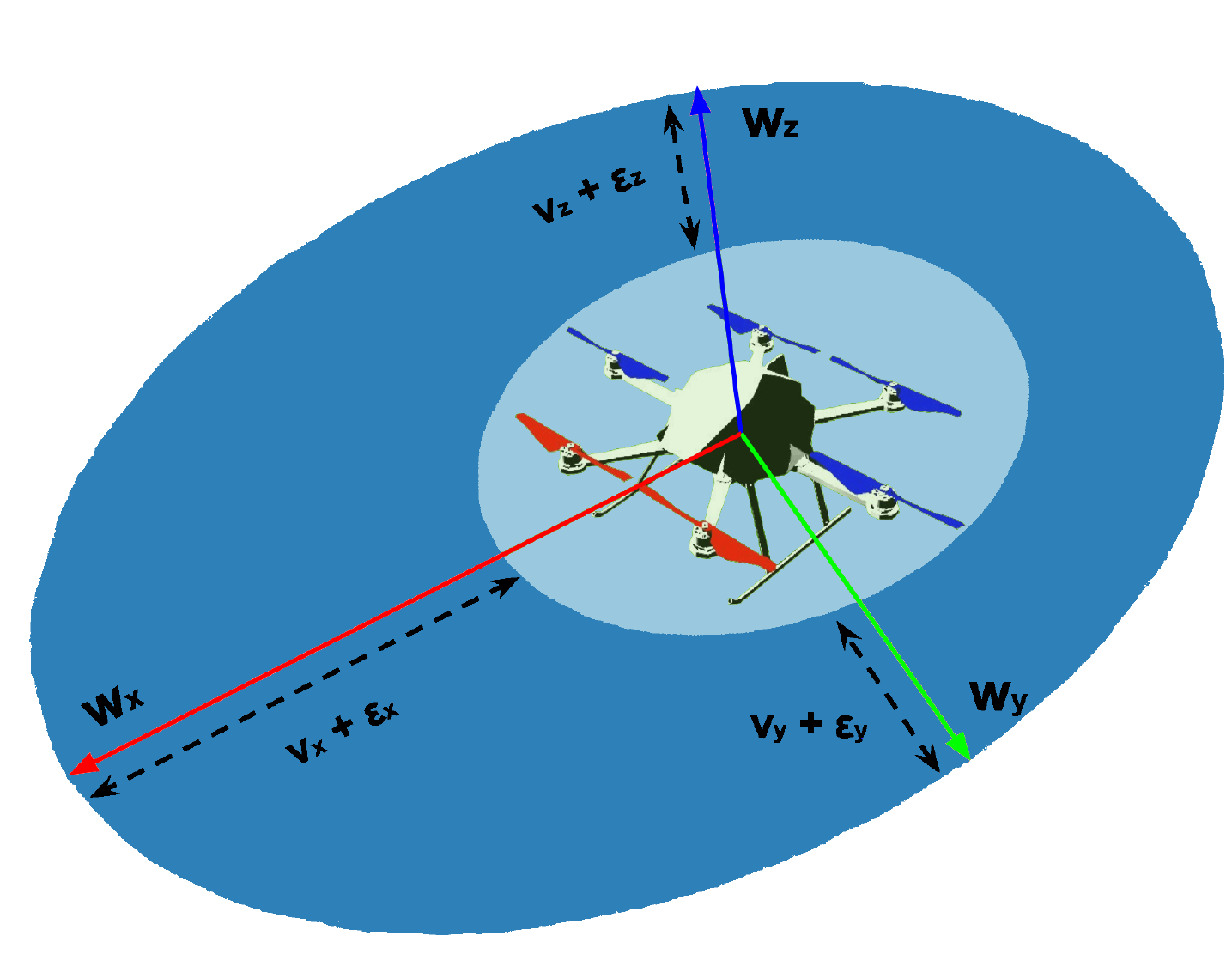}
 \caption{Ellipsoidal bounding box concept overview: the light blue area bounds the obstacle physical dimensions, while the blue area embeds the uncertainties $\epsilon$ in the obstacle pose estimation and velocity $v$. The blue area is stretched along the x axis direction to take into account the object estimated velocity.}
\label{fig:ellipsoid}
 \end{figure}

Let $o_{WO_i} =(o_{x_i}, o_{y_i}, o_{z_i})^T$ be the 3D position of the \textit{i-th} obstacle in the world frame $W$. To enable the UAV to safely flight, the trajectory has to constantly keep the aerial vehicle at a safe distance from all the surrounding obstacles. Moreover, the cost function\eqref{eq:cost_function} has to take into account objects with different shapes and sizes. Thus, we formulate the avoidance cost $c_o(x_t,o_t,u_t)$ as:
\begin{align}
   \sum_{i=1}^{N_o}\frac{1}{d_iW_id_i^T}, \hspace{0.5cm}&d_i = P_{WB} - o_{WO_i} \\\nonumber 
   W_i = diag(w_{xi}, w_{yi}, &w_{zi}),\hspace{0.2cm}w_i= f(\gamma_i, \epsilon_i, v_i) 
\end{align}
where $N_o$ is the number of the obstacles and $W_i$ is the i-th weighting matrix. The latter weighs the distances along the 3 main axes, creating an ellipsoidal bounding box. More specifically, each component $w_i$ embeds  the obstacle's size $\gamma$, velocity $v$, and estimation uncertainties $\epsilon$ (see \figref{fig:ellipsoid}). Among the others, this formulation allows to set more conservative bounding boxes according to the obstacle detection accuracy. Moreover, to guarantee a robust collision avoidance, we formulate the minimum acceptable distance as an additional inequality $h(x_t, o_t, u_t)$ constraint:  
\begin{align}
  \sum_{i=1}^{N_o}d_iW_id_i^T >= d_{min,i}
\end{align}
where $d_{min,i}$ represents the minimum acceptable distance for the i-th obstacle.

\subsection{Action Objectives}
\label{sec:action_obj}

The action objectives act to penalize the amount of control inputs used to steer the vehicle. Therefore, we formulate the action cost $c_a(x_t, u_t)$ as:
\begin{align}
 c_a(x_t, u_t) = \bar{u} R \bar{u}^T, \hspace{.5cm}\bar{u} = u - u_{ref}
\end{align}
where $R$ is a weighting matrix, and $u_{ref}$ represents the reference control input vector (e.g. the control commands to keep the aerial vehicle in hovering). Moreover, to constrain 
the control commands to be bounded inside the input range allowed by the real system, we add an additional inequality constraint $h(x_t, u_t)$: 
\begin{align}
   u_{lb} <= u <= u_{ub}
\end{align}

The remaining costs $c_n(x_t, u_t)$ and $c_f(x_{tf})$ penalize the distance from the goal pose, and are formulated as:

\begin{align}
 c_n(x_t, u_t) = \bar{x} Q \bar{x}^T&, \hspace{.5cm}\bar{x} = x - x_{ref}\nonumber\\
 c_f(x_{tf}) = \bar{x}_{tf} Q_N \bar{x}_{tf}^T, &\hspace{.5cm}\bar{x}_{tf} = x_{tf} - x_{ref}
\end{align}

\begin{figure*}[ht]
 \centering
 \begin{subfigure}{0.65\columnwidth}
        \includegraphics[width=\columnwidth]{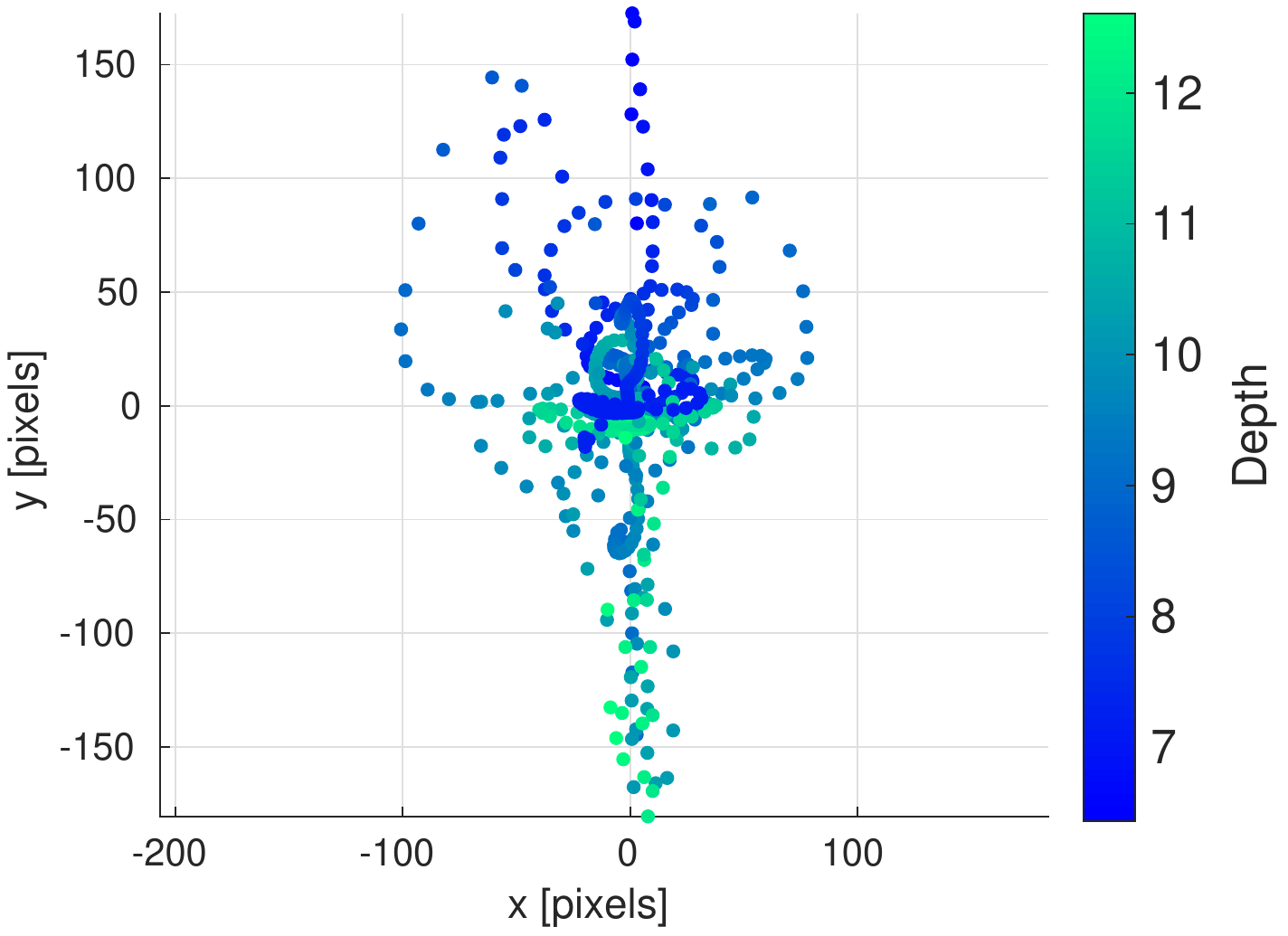}
        \caption{Target depth color map}\label{fig:repr_err_statica}
 \end{subfigure}
 \begin{subfigure}{0.65\columnwidth}
        \includegraphics[width=\columnwidth]{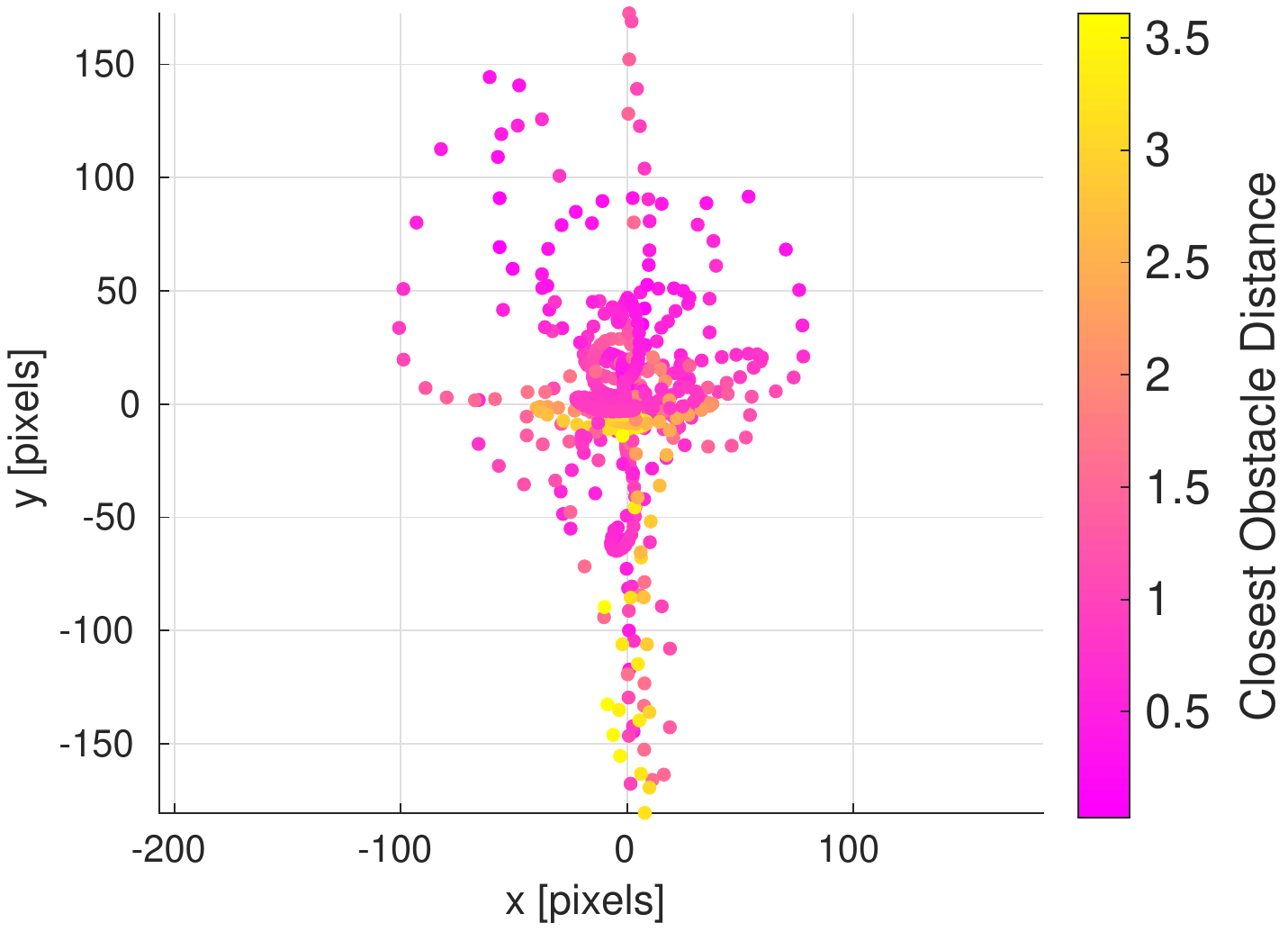}
        \caption{Closest obstacle distance color map}\label{fig:repr_err_staticb}
 \end{subfigure}
 \begin{subfigure}{0.65\columnwidth}
        \includegraphics[width=\columnwidth]{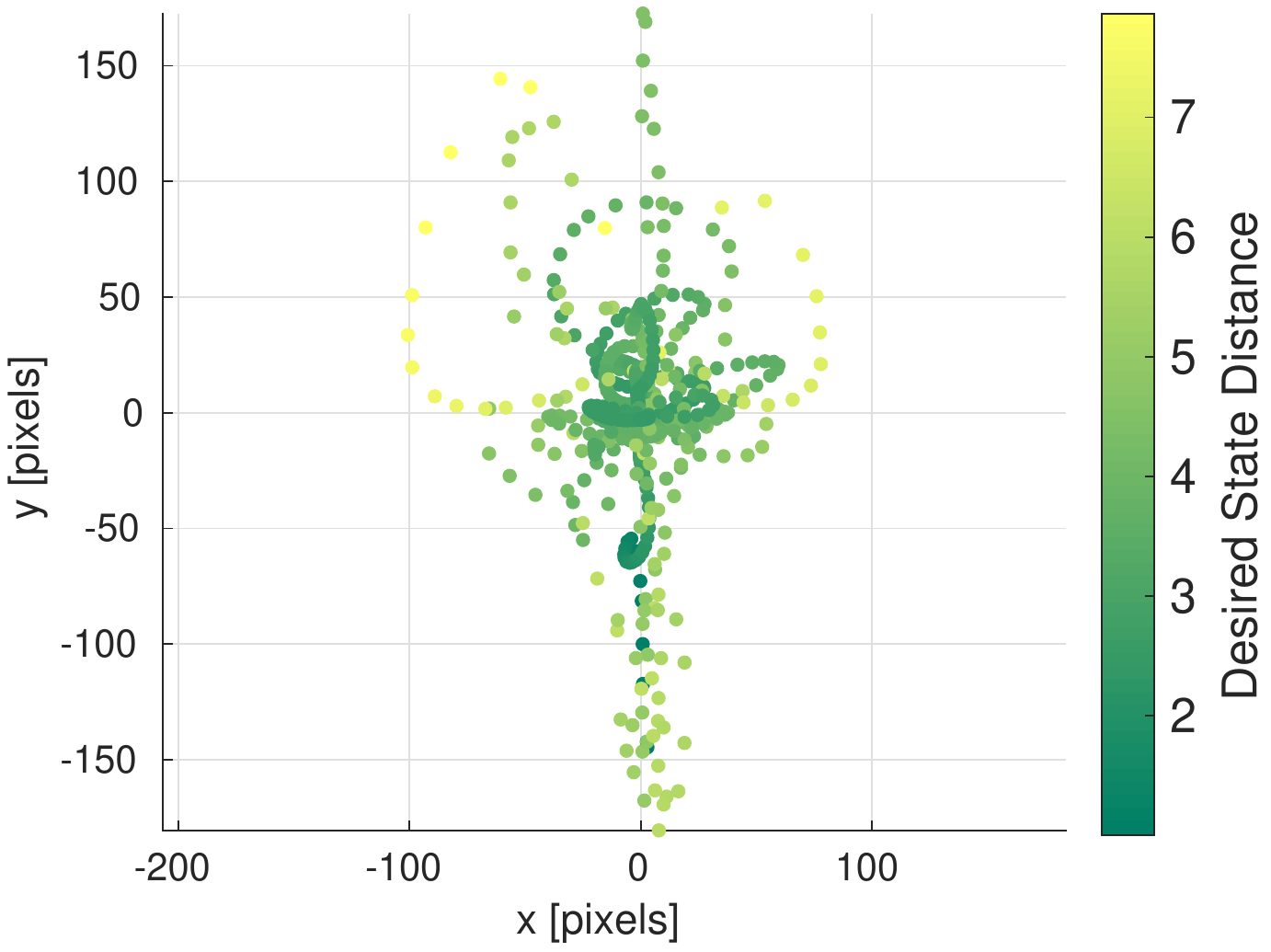}
        \caption{Desired pose distance color map}\label{fig:repr_err_staticc}
 \end{subfigure}
 \caption{Target reprojection error for 10 hover-to-hover flights. The three different color-maps represent the depth of the point of interest, the distance from the closest static obstacle, and the distance between the current pose and the desired pose, respectively. }
\label{fig:repr_err_static}
 \end{figure*}

\section{Non-Linear Model Predictive Control}
\label{sec:overview}

The cost function given in\eqref{eq:cost_function} results in a non-linear optimal control problem. To find a time-varying control law that minimizes it, we make use of a Non-Linear Model Predictive Controller, where the cost function\eqref{eq:cost_function} is firstly approximated by a Sequential Quadratic Program (SQP), and then iteratively solved by a standard Quadratic Programming (QP) solver.

The whole system works in a reciding horizon fashion, meaning that at each new measurement, the NMPC provides a feasible solution and only the first control input of the provided trajectory is actually applied to control the robot. 

To achieve that, we discretize the system dynamics with a fixed time step $dt$ over a time horizon $T_H$ into a set of state vectors $x_{0:N} = \{x_0, x_1, \dots, x_{N}\}$ and a set of inputs controls $u_{0:N} = \{u_0, u_1, \dots, u_{N-1}\}$, where $N = T_H/dt$. We also define the state, the final state, and input cost matrices as $Q$, $Q_N$, and $R$, respectively. The final cost function will be:
\begin{align}
 J = \bar{x}_{tf} Q_N \bar{x}_{tf}^T + \sum_{i=0}^{N-1} (c_n + c_a + c_p + c_o)
\end{align}
where $\bar{x}$ represents the difference with respect to the state reference values, while $c_n$, $c_a$, $c_p$ and $c_o$ refer to the  navigation, action, perception and avoidance objectives introduced in the previous section. 

For NMPC to be effective, the optimization should be performed in real-time. In this regard, we compute an approximation of each optimal solution by executing only a few iterations at each control loop. Moreover, we keep the previous approximated solution as the initialization for the next optimization.

\section{Experiments}
\label{sec:exp}

The evaluation presented here is designed to support the claims made in the introduction. 
We performed two kinds of experiments, namely \textit{hover-to-hover flight with static obstacles} and \textit{hover-to-hover flight with dynamic obstacles}. To demonstrate the real-time capabilities of the proposed approach we also report a computational time analysis. In all the reported results, the multirotor is asked to fly multiple times by randomly changing the obstacles setup and the goal state.

\subsection{Simulation Setup}
We tested the proposed approach in a simulated environment by using the RotorS UAV simulator \cite{furrer2016springer} and an AscTec Firefly multirotor. We setup the non-linear control problem with the ACADO toolbox and used the qpOASES solver \cite{Ferreau2006phd}. By using the ACADO code generation tool, the problem is then exported in a highly efficient C code that we integrated within a ROS (Robot Operating System) node. We set the discretization step to be $dt = 0.2~s$ with a time horizon $T_H = 2~s$. To guarantee enough agility to the vehicle, we run the control loop at $100~Hz$. The mapping between the optimal control inputs and the propeller velocities is done by a low-level PD controller that aims to resemble the low-level controller that runs on a real multirotor. To make the simulation more realistic, we add a further white noise with standard deviation $\sigma$ on the 3D positions of the detected obstacles. The code developed in this work is publicly available as open-source software. 

\begin{figure*}[ht]
\setlength{\tabcolsep}{1pt}
\begin{tabular}{ccc}
\begin{subfigure}{0.25\textwidth}\centering\includegraphics[width=45mm]{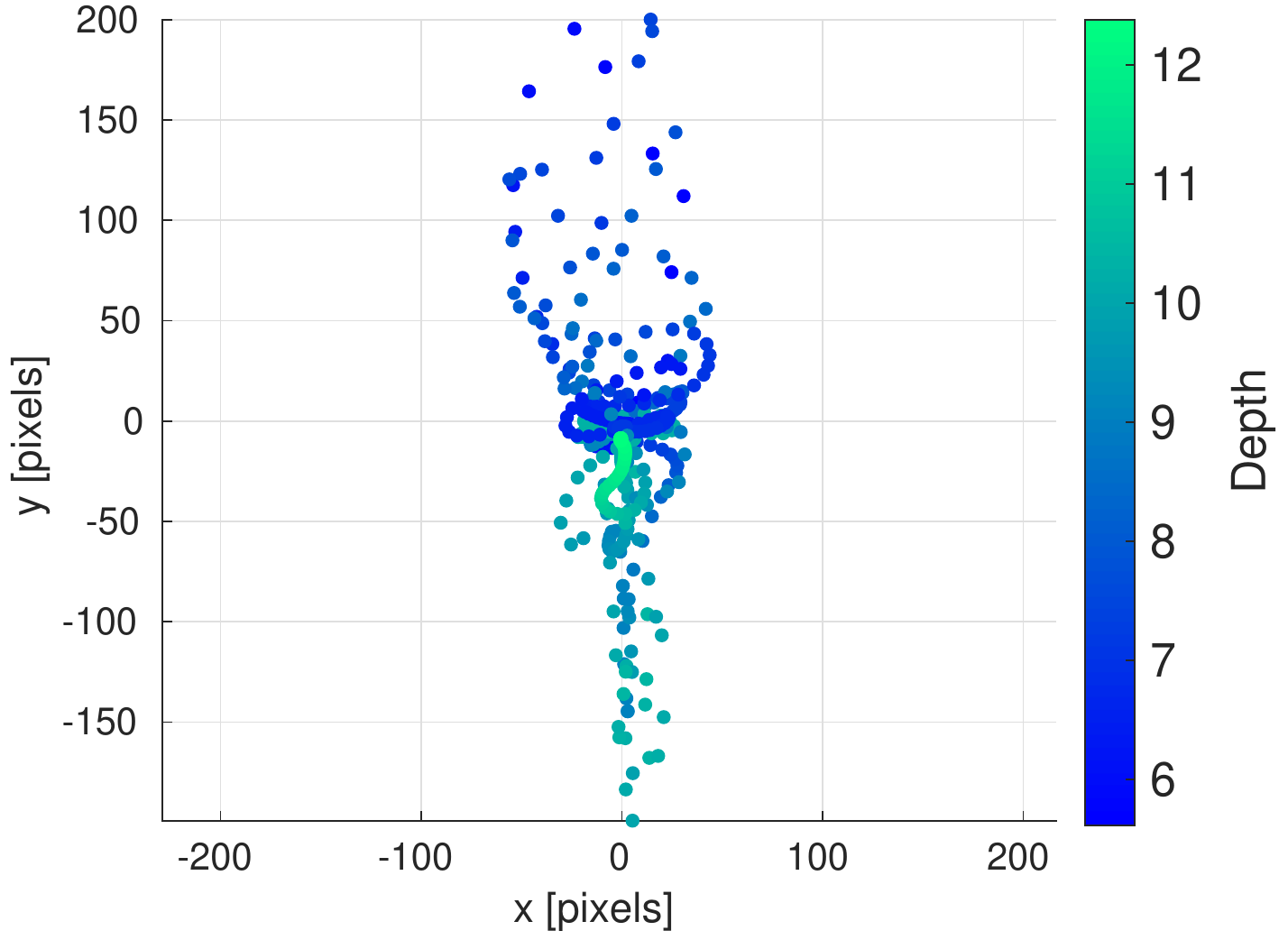}\caption{Target depth color map}\label{fig:taba}\end{subfigure}&
\begin{subfigure}{0.25\textwidth}\centering\includegraphics[width=45mm]{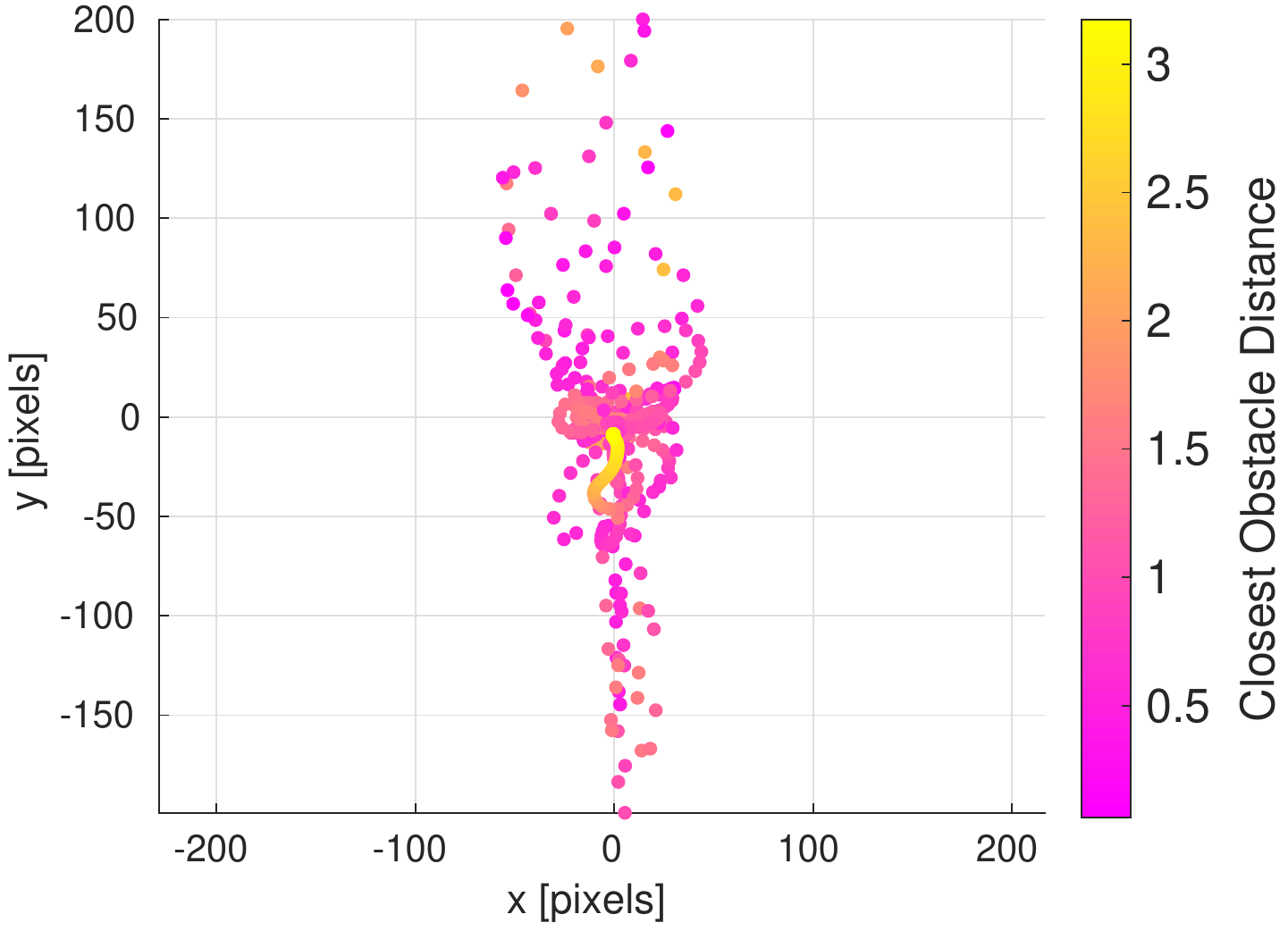}\caption{Closest obstacle dist. color map}\label{fig:taba}\end{subfigure}&
\begin{subfigure}{0.5\textwidth}\centering\includegraphics[width=80mm]{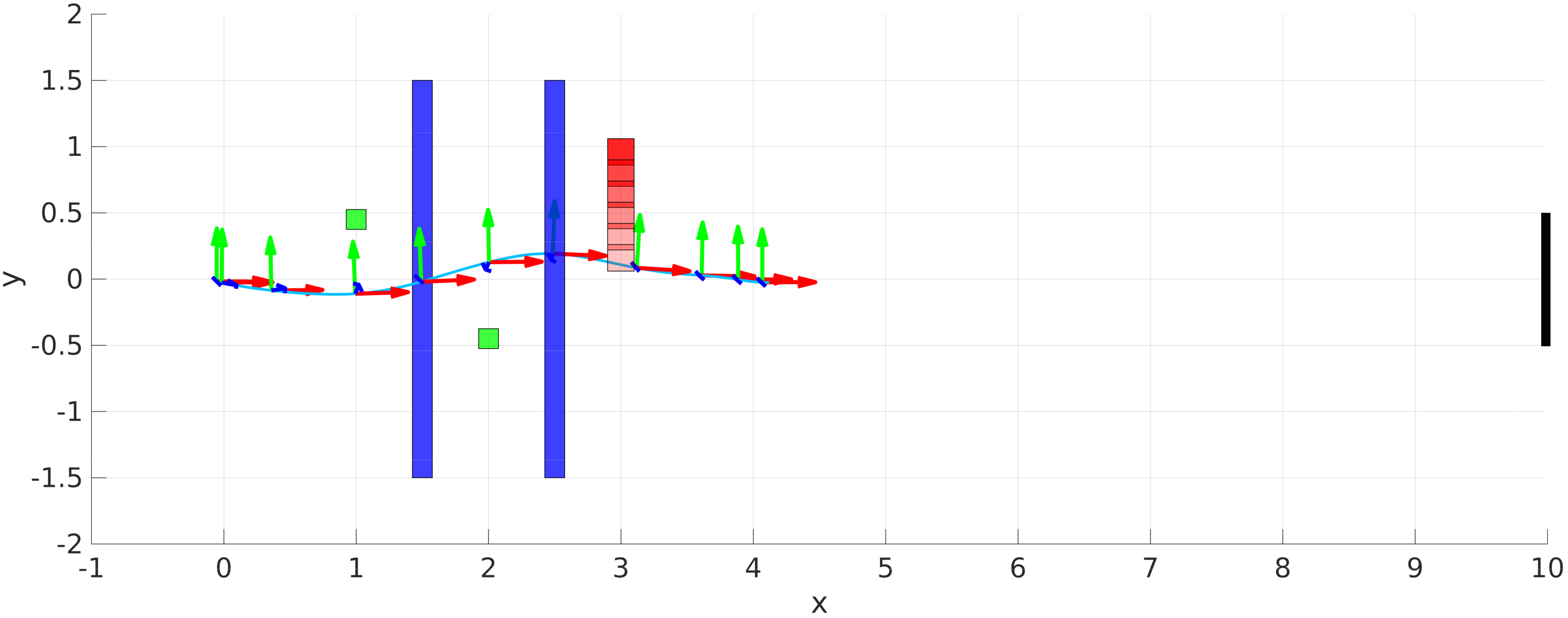}\caption{Example trajectory top view}\label{fig:tabe}\end{subfigure}\\
\newline
\begin{subfigure}{0.25\textwidth}\centering\includegraphics[width=45mm]{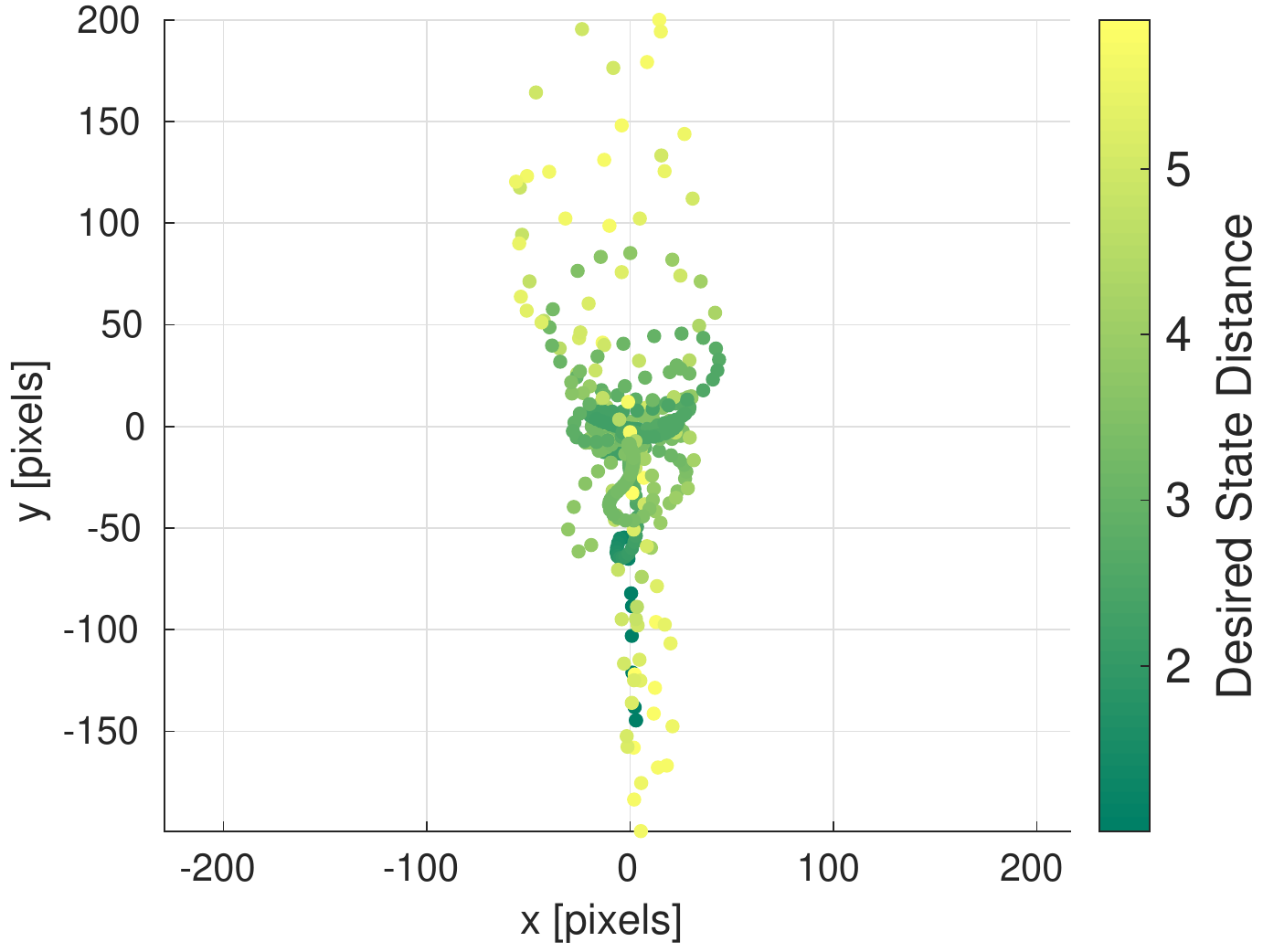}\caption{Desired pose dist. color map}\label{fig:tabc}\end{subfigure}&
\begin{subfigure}{0.25\textwidth}\centering\includegraphics[width=45mm]{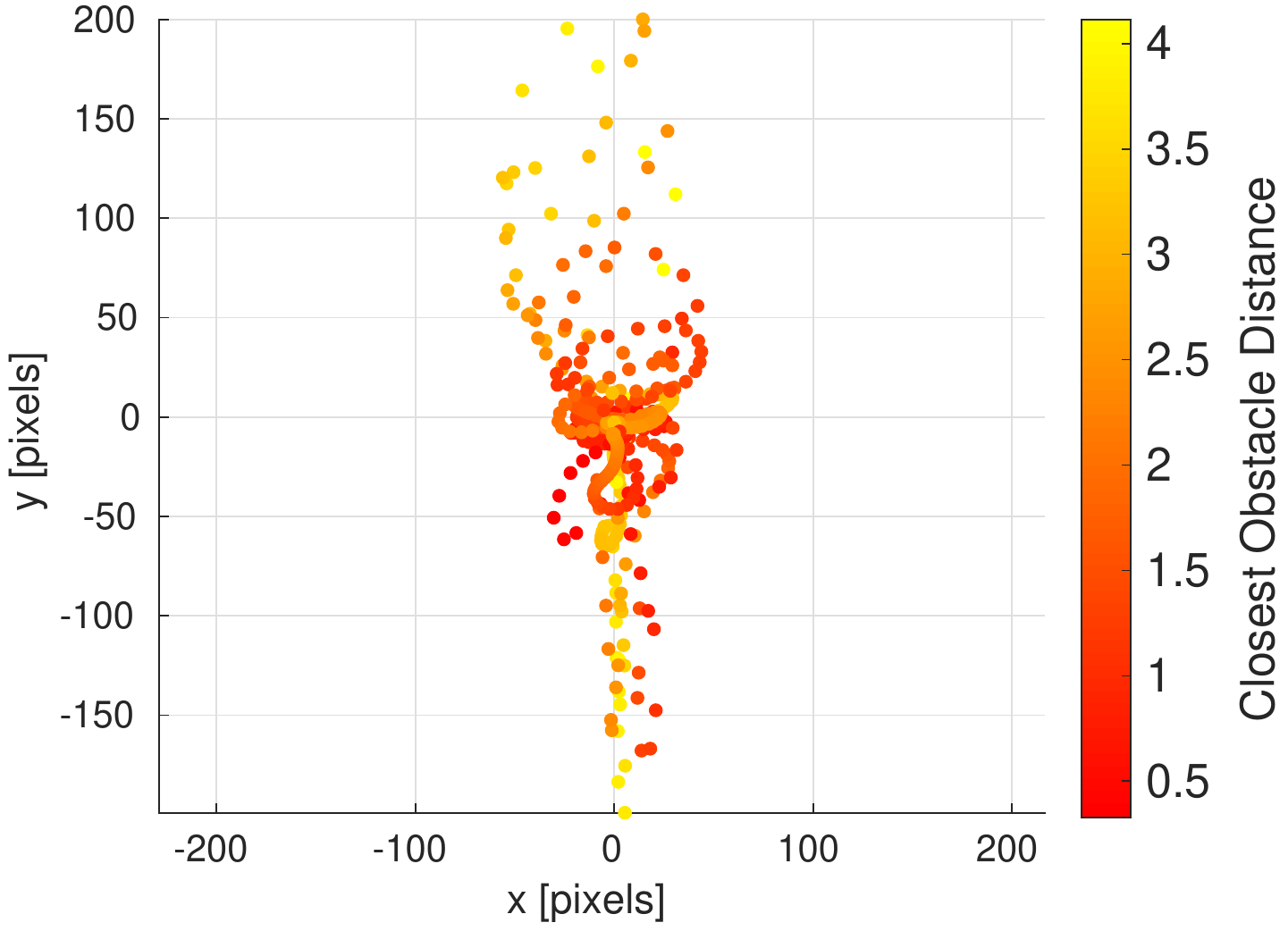}\caption{Dyn. obstacle dist. color map}\label{fig:tabd}\end{subfigure}&
\begin{subfigure}{0.5\textwidth}\centering\includegraphics[width=80mm]{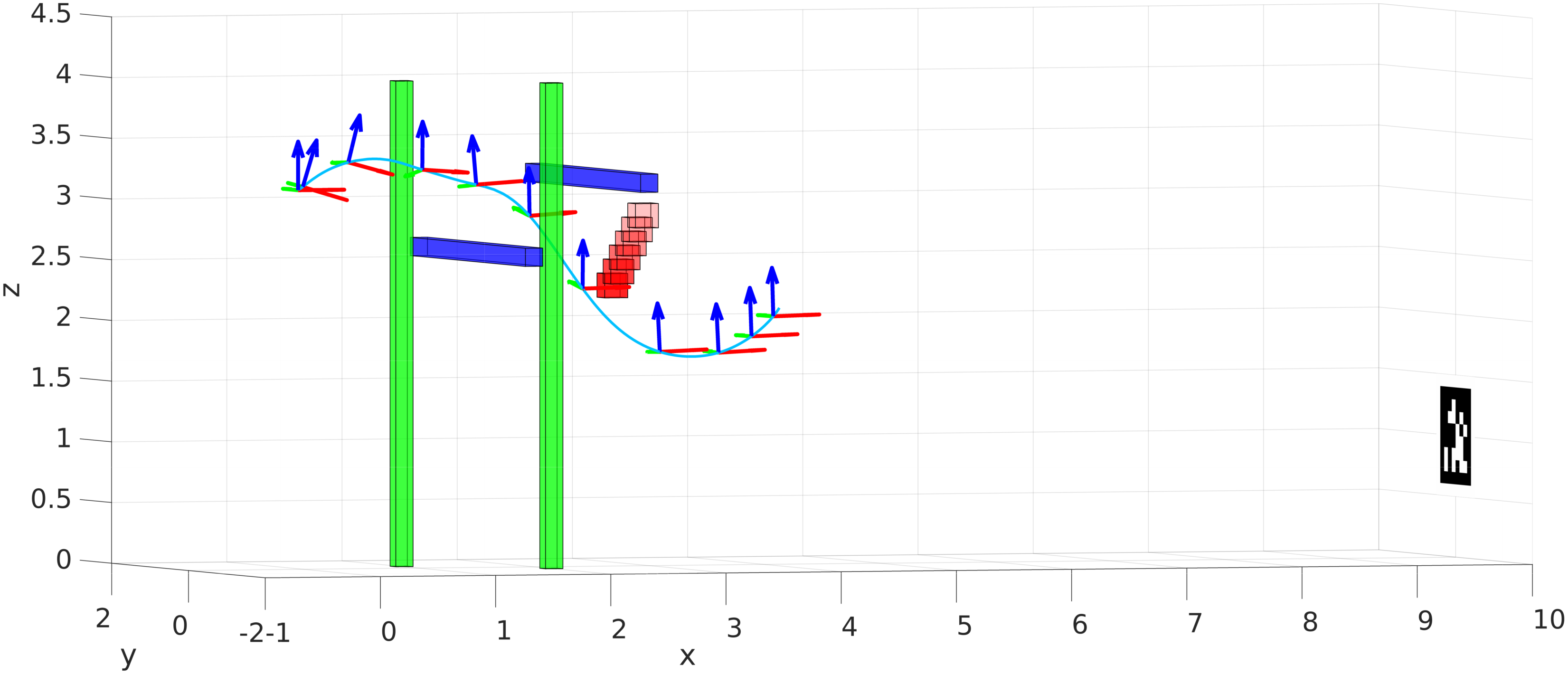}\caption{Example trajectory lateral view}\label{fig:tabf}\end{subfigure}\\
\end{tabular}
\caption{Reprojection error in the hover-to-hover flight with dynamic obstacles (left-side), and an example of planned trajectory (right-side). The moving, dynamic, obstacle is represented by the red cube where the varying color intensity represents its motion over the time.}
\label{fig:dyn_flight}
\end{figure*}

\subsection{Hover-To-Hover Flight with Static Obstacles}
\label{ex:hov_static_obst}

\begin{table}[t!]

    \scriptsize
	\centering
    \setlength{\tabcolsep}{1.5pt} 
    
	\begin{tabular}{ cccccccc }
	   \makecell{Dynamic \\ Obstacle} &delay [s] &\makecell{failure \\ rate [\%]}  &\makecell{ Avg. pixel \\ error} 
      &\makecell{Max pixel \\ error} &Torque [N] &$T_{cmd} [g]$ &$\sigma[cm]$ \\\cline{1-8}\noalign{\vskip 1mm}
	             &      &$0.2$   &$53$  &$98$   &0.029  &1.35 &2 \\
	  \checkmark &$.2$  &$0.4$   &$79$  &$129$  &0.030  &1.39 &2 \\
	  \checkmark &$.4$  &$2.0$   &$84$  &$142$  &0.031  &1.41 &2 \\
	  \checkmark &$.6$  &$10.4$  &$90$  &$151$  &0.040  &1.47 &2 \\
	  \checkmark &$.8$  &$18.9$  &$92$  &$155$  &0.040  &1.50 &2 \\
	  \checkmark &$1$   &$24.8$  &$98$  &$157$  &0.041  &1.52 &2 \\
	      
	  \end{tabular}	
        \captionof{table}{Trajectory statistics comparison across different simulation setups.}
	\label{tab:simul_statistics}
    
    \vspace{0.1cm}

    \begin{tabular}{ cccccccc }
	   \makecell{Dyn. obst.\\delay} &\makecell{Dyn. obst.\\velocity} [s] &\makecell{failure \\ rate [\%]}  &\makecell{ Avg. pix. \\ error} 
      &\makecell{Max pix. \\ error} &Torque [N] &$T_{cmd} [g]$ &$\sigma[cm]$ \\\cline{1-8}\noalign{\vskip 1mm}
	  $.2$  &$.2$  &$1.9$  &$81$   &$101$  &0.031  &1.40 &2 \\
	  $.2$  &$.4$  &$3.5$  &$89$   &$111$  &0.031  &1.41 &2 \\
	  $.2$  &$.6$  &$4.8$  &$90$   &$113$  &0.034  &1.44 &2 \\
	  $.4$  &$.2$  &$3.9$  &$93$   &$140$  &0.032  &1.45 &2 \\
	  $.4$  &$.4$  &$7.4$  &$94$   &$147$  &0.034  &1.44 &2 \\
	  $.4$  &$.6$  &$11.9$ &$107$  &$151$  &0.035  &1.49 &2 \\
	      
	  \end{tabular}	
        \captionof{table}{Trajectory statistics comparison across different dynamic obstacle spawning setups.}
    \label{tab:spawn_statistics}

\end{table}

In this experiment, we show the capabilities in \textit{hover-to-hover} flight maneuvers with static obstacles. More specifically, the UAV is commanded to reach a set of $M$ randomly generated desired states $X_{des} = (x_{ref,0}, x_{ref,1}, \dots, x_{ref,M} )$. Unlike standard controllers, the proposed approach will generate, at each time step, control inputs that will steer the vehicle towards the goal state while avoiding obstacles and keeping the target in the image plane. \figref{fig:repr_err_static} depicts the reprojection error of the point of interest and its correlation with (i) the depth of the point of interest, (ii) the distance from the closest obstacle, and (iii) the distance from the desired state. The largest reprojection errors occur when the UAV is farther from the desired state, or when the UAV has to fly closer to the obstacles. In these cases, the reprojection error is slightly higher since the UAV has to perform more aggressive maneuvers. However, as reported in Tab.~\ref{tab:simul_statistics}, the UAV keeps a success rate of almost $100\%$ while keeping a low usage of control inputs.
\begin{figure}[t!]
 \includegraphics[width=\columnwidth]{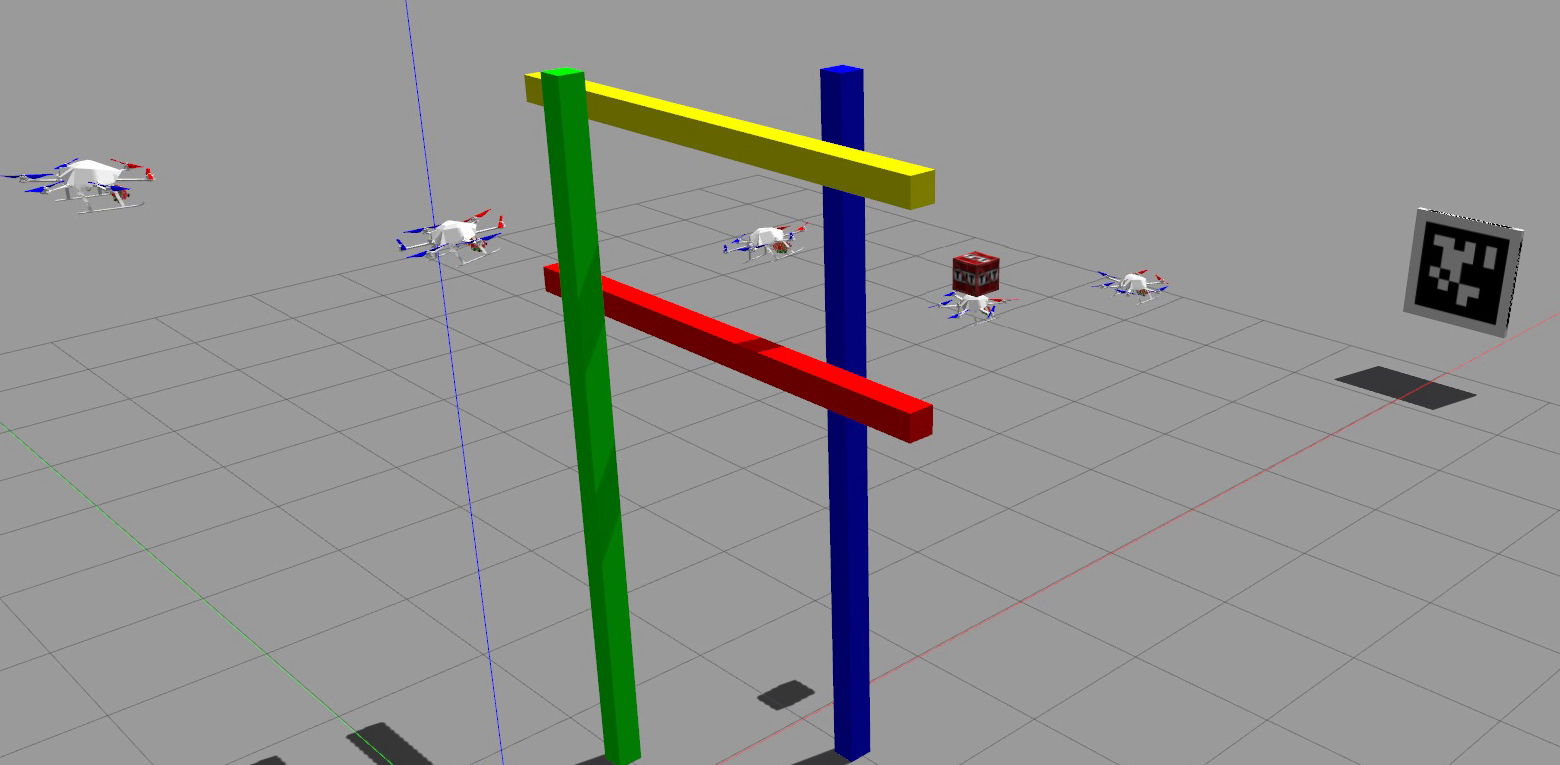}
 \caption{Example of a trajectory generated in our simulated scenario. The depicted UAVs represent different poses assumed by the aerial vehicle across the time horizon. The colored objects represent the static obstacle, while the red box the dynamic one.}
\label{fig:timelapse}
 \end{figure}

\subsection{Hover-To-Hover Flight with Dynamic Obstacles}
\label{ex:hov_dynamic_obst}

This experiment shows the capabilities to handle more challenging flight situations, such as the flight in presence of dynamic, unmodeled obstacles (see \figref{fig:timelapse} for an example). To demonstrate the performance in such a scenario, we randomly spawn a dynamic obstacle along the planned trajectory. Thus, to successfully reach the desired goal, the UAV has to quickly re-plan a safe trajectory (see \figref{fig:tabc} and \figref{fig:tabf} for an example). Moreover, to make experiments with an increasing level of difficulty, we spawn the dynamic obstacle with a random delay and with a random non-zero velocity. The random delay simulates the delay in detecting the obstacle, or the possibility that the obstacle appears after the vehicle has already planned the trajectory. 

\begin{figure}[t!]
          \includegraphics[width=.95\columnwidth]{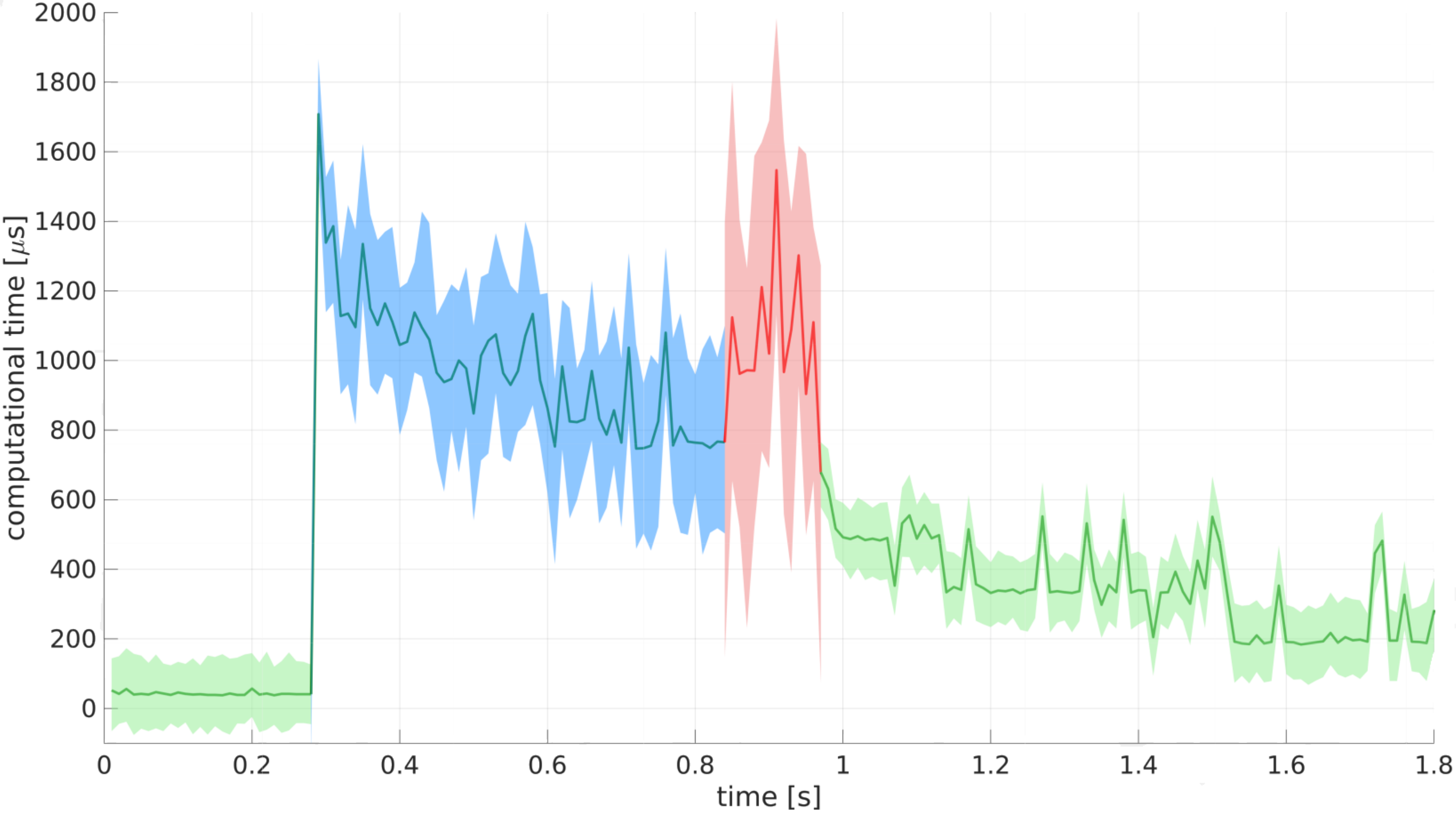}
          \caption{Average computational time plot across the planning phases: the (i) planning phase in blue, (ii) the steady-planning phase in green, and (iii) the emergency re-planning phase in red. The shaded areas represent the variance of the average computational time.}
          \label{fig:run_time_plot}
\end{figure}

The reprojection error follows a similar trend as the previous set of experiments (see \figref{fig:dyn_flight}), being higher when the vehicle is closer to static obstacles of farther from the desired state. In \figref{fig:tabd} we also report the evolution of the reprojection error colored according to the distance from the dynamic obstacle. Since the latter is spawned close to the planned trajectory, the UAV has to perform an aggressive maneuver to keep a safe distance from it. This usually leads to have a smaller reprojection error when the dynamic obstacle is close (i.e. the object is spawned while the UAV was on the optimal trajectory), and a bigger error when the obstacle is farther (i.e. the drone reacted with an aggressive maneuver to avoid it). Tab.~\ref{tab:simul_statistics} and Tab.~\ref{tab:spawn_statistics} report some trajectory statistics. The proposed method keeps a success rate above the $75\%$ in almost all the conditions, even in presence of large delays. It is also noteworthy to highlight how the delay turns out to be more critical than the dynamic obstacle's velocity. The latter, indeed, makes the re-planning more challenging only in specific circumstances (e.g. when the object moves towards the UAV). Finally, the capability to avoid obstacles comes with a performance trade-off. The greater the difficulty, the greater the use of control inputs. This is especially true when the UAV has to avoid dynamic obstacles with a large delay, since it involves making expensive control maneuvers.

\subsection{Computational Time}
\label{ex:comp_time}

To meet the control loop real-time constraints, the NMPC computational cost should be as low as possible. Moreover, the computational cost is not constant, and might vary according to the similarity between the initial trajectory and the optimal one. In this regards, we distinguish among three main flight phases: 
\begin{enumerate}
 \item the planning phase, which occurs when the UAV has to plan a trajectory to reach $x_{des}$;
 \item the steady-planning phase, which occurs when the UAV is already moving toward $x_{des}$;
 \item the emergency re-planning, which occurs when a dynamic obstacle suddenly appears along the optimal trajectory.
\end{enumerate}
\figref{fig:run_time_plot} reports the average computational costs for all those flight phases. The average computational cost is constantly lower than  0.01 seconds, meeting the control loop frequency constraints. The steady-planning phase turns out to be the cheapest one. Indeed, since the control loop runs 100 times per second, the neighbour trajectories are quite similar. Conversely, the emergency re-planning phase is the most expensive and variable, since the trajectory to re-plan is often quite different from the previous one, depending on where the dynamic obstacle is spawned. 

\section{Conclusion}
\label{sec:conclusion}

This work proposes an NMPC controller for enhancing vision-based navigation with static and dynamic obstacle avoidance capabilities. The proposed method represents the obstacles with customly shaped constraints by taking into account their velocity and uncertainties, and making it possible to adapt the safety of the planned trajectory. The capabilities of this system have been extensively tested in a simulated environment, and across different scenarios. The experiments suggest that the proposed method allows to safely fly even in challenging situations. We release our C++ open-source implementation,
enabling the research community to test the proposed algorithm. Future work will investigate the performance of the proposed system in real-world scenarios.

\balance

\bibliographystyle{plain}
\bibliography{glorified.bib}

\end{document}